%% file: Paper-105.tex
\documentclass[runningheads]{llncs}
\usepackage[T1]{fontenc}
\usepackage{graphicx,verbatim}
\usepackage{booktabs} % \toprule, \midrule, \bottomrule
\usepackage{multirow} % \multirow
\usepackage{adjustbox} % to enable too wide tables
\usepackage{rotating}
\usepackage{tabularx}
\usepackage{float}

\usepackage{pifont}
\usepackage{pgfplots} % To handle pgf figures, allows LaTeX fonts & vector graphics
\usepackage{tikz}
\usepackage{tikz-3dplot}
\usetikzlibrary{positioning}
\pgfplotsset{compat=1.18}
\usetikzlibrary{3d,calc}

\graphicspath{{reconstruction}} % this is the place where all rasterized images (i.e. png that are part of some .pgf figure) are stored and loaded from. 
\usepgfplotslibrary{external} 
\definecolor{ourdarkorange}{HTML}{CC8933}

\definecolor{ourorange}{HTML}{FFD59F}
\definecolor{ourblue}{HTML}{6D9EEB}
\definecolor{ourred}{HTML}{E06666}
\definecolor{ouryellow}{HTML}{FFD966}
\definecolor{ourgreen}{HTML}{93C47D}

\definecolor{ourlightblue}{HTML}{CFE2F3}
\definecolor{ourlightred}{HTML}{F4CCCC}
\definecolor{ourlightyellow}{HTML}{FFF2CC}
\definecolor{ourlightgreen}{HTML}{D9EAD3}

\usepackage{amsmath, amssymb}
\usepackage{hyperref}

\begin{document}
\title{MultiMAE for Brain MRIs: Robustness to Missing Inputs Using Multi-Modal Masked Autoencoder}

\titlerunning{Multi
MAE for Brain MRIs}

\author{%
Ayhan Can Erdur\inst{1,2,\ast} \and
Christian Beischl\inst{2,\ast} \and
Daniel Scholz\inst{2,3,4} \and
Jiazhen Pan\inst{2} \and
Benedikt Wiestler\inst{3,4, \bullet} \and
Daniel Rueckert\inst{2,4,5,\bullet} \and 
Jan C. Peeken\inst{1,6,7,\bullet}
}

\authorrunning{Erdur \& Beischl et al.}



\institute{%
Department of Radiation Oncology, TUM University Hospital, Munich, Germany \and
Chair for AI in Healthcare and Medicine, Technical University of Munich (TUM) and TUM University Hospital, Munich, Germany \and
Chair for AI for Image-Guided Diagnosis and Therapy, Technical University of Munich (TUM) and TUM University Hospital, Munich, Germany \and
Munich Center for Machine Learning (MCML), Munich, Germany \and
Department of Computing, Imperial College London, London, UK \and
Deutsches Konsortium für Translationale Krebsforschung (DKTK), Partner Site Munich, Munich, Germany \and
Institute of Radiation Medicine (IRM), Department of Radiation Sciences (DRS), Helmholtz Center Munich, Munich, Germany \\
\textsuperscript{*} contributed equally as first authors, \textsuperscript{$\bullet$} contributed equally as last authors  \\
 \email{can.erdur@tum.de}
}    
\maketitle              % typeset the header of the contribution
\begin{abstract}
Missing input sequences are common in medical imaging data, posing a challenge for deep learning models reliant on complete input data. In this work, inspired by MultiMAE \cite{bachmann2022multimae}, we develop a masked autoencoder (MAE) paradigm for multi-modal, multi-task learning in 3D medical imaging with brain MRIs. Our method treats each MRI sequence as a separate input modality, leveraging a late-fusion-style transformer encoder to integrate multi-sequence information (“multi-modal”) and individual decoder streams for each modality for “multi-task” reconstruction. This pretraining strategy guides the model to learn rich representations per modality while also equipping it to handle missing inputs through cross-sequence reasoning. The result is a flexible and generalizable encoder for brain MRIs that infers missing sequences from available inputs and can be adapted to various downstream applications. We demonstrate the performance and robustness of our method against an MAE-ViT baseline in downstream segmentation and classification tasks, showing absolute improvement of $10.1$ overall Dice score and $0.46$ MCC over the baselines with missing input sequences. Our experiments demonstrate the strength of this pretraining strategy. The implementation is made available \footnote{The repository for our work: \url{https://github.com/chris-beischl/multimae-for-brain-mri}}.

\keywords{Masked Autoencoder Pretraining  \and Brain MRI \and Missing Modalities}
% Authors must provide keywords and are not allowed to remove this Keyword section.

\end{abstract}

\section{Introduction}
% to introduce: 
% - ViTs, self-supervised learning for pretraining
% - both in medical imaging
% - we adapt MultiMAE to medical 
% - introduce downstream tasks (especially label names GBM, Astro, Oligo)
% - missing modality use cases

% would be nice to have the last paragraph as a list of contributions (see Daniel's MM-DINO)
Pretraining has become a cornerstone of modern deep learning, particularly in data-scarce domains like medical imaging, where labeled annotations are costly and limited.
Self-supervised learning offers an effective solution, enabling models to learn transferable representations from unlabeled scans and consistently outperform training from scratch \cite{azizi2021big,shurrab2022self,zhou2023maeunetr}. The basis for the self-supervised learning strategy is the use of a pretext task, such as the reconstruction of masked parts of the inputs \cite{he2022masked}. 

Transformer-based architectures have rapidly gained traction in medical imaging, both as pure Vision Transformers (ViTs) \cite{dosovitskiy2021vit} and in hybrid CNN-transformer designs, showing strong performance in classification \cite{dai2021transmed,manzari2023medvit}, and segmentation \cite{hatamizadeh2022unetr,jiang2022swinbts}.
% and more \cite{shao2023hvtsurv,erdur2025improving}.

Despite architectural differences, there is one common approach adopted by most of the existing work with magnetic resonance imaging (MRI): processing multiple modalities by stacking them along the channel dimension, analogous to RGB channels in natural images \cite{hatamizadeh2022unetr,jiang2022swinbts}. 
While this early-fusion strategy is simple, it requires all modalities to be present during training and inference, which very often is not the case in real-life clinical environments. 
Moreover, it limits modality-specific representation learning and leads to sharp performance drops when inputs are missing \cite{zhou2023literature}.

In this work, we adopt a MultiMAE-based \cite{bachmann2022multimae} pretraining framework with modality-specific encoding to strengthen the learning of distinct representations per sequence and their cross-modal integration, rather than treating all inputs as a single fused whole from the outset, and masked modeling to improve robustness to missing modalities. 
This approach introduces only minor changes to the ViT backbone and outperforms standard MAE \cite{he2022masked} pretraining in both segmentation and classification, particularly under missing-modality conditions.
Our key contributions are as follows:

\begin{enumerate}
    \item We investigate MultiMAE pretraining for 3D brain MRI, enabling modality-specific encoding and masked modeling across multiple MRI sequences.
    \item We show that \textbf{ pretraining improves performance and robustness to missing input modalities} in both segmentation (glioma, metastasis) and classification of glioma subtypes into glioblastoma (GBM), astrocytoma (Astro), and oligodendroglioma (Oligo) according to WHO 2021 standards. The improvement is evident both in internal and external test sets.
    \item We demonstrate that MultiMAE can \textbf{synthesize entirely missing modalities} at inference time.
\end{enumerate}

\section{Related Work}

\textbf{Self-Supervised and Multi-Modal Learning in Medical Imaging}
Medical image analysis often suffers from scarce annotated data due to the high cost and effort of expert labeling \cite{tang2022self,azizi2021big}. Self-supervised learning (SSL) addresses this by leveraging large unlabeled datasets to boost downstream task performance, a strategy increasingly adopted in medical imaging \cite{zhou2019models,tang2022self}.

SSL approaches include generative methods like masked autoencoding \cite{zhou2023maeunetr}, discriminative contrastive learning \cite{azizi2021big,chaitanya2020contrastive}, and hybrid objectives combining both \cite{zhou2019models,tang2022self}. Masked autoencoders (MAEs) \cite{he2022masked} have been extended to 3D medical data, with MAE-UNETR \cite{zhou2023maeunetr} demonstrating notable gains in segmentation. 

\noindent\textbf{Handling Missing Modalities and Robustness}
Multi-modal learning leverages complementary data sources, such as images and text \cite{wang2022medclip}, or different imaging acquisitions. Typically, models use modality-specific encoders with fusion layers to capture cross-modal interactions. Besides performance benefits, such fusion improves robustness to missing modalities \cite{zhang2022mmformer,wang2023shaspec}.

For natural images, MultiMAE \cite{bachmann2022multimae} offers a simple and versatile multi-modal SSL to learn joint representations across imaging modalities. It remains, however, underexplored in medical imaging.

% \textbf{Multi-Modal and Self-Supervised Learning in Medical Imaging} Masked autoencoders (MAEs) \cite{he2022masked} have shown strong performance in self-supervised learning for medical imaging, where access to large-scale data is limited. 
% %maybe first some general work that underlines improvement by MAE, then MAE-UNETR (and one more) segmentation, then multi-modal approaches for 2D?, shortly going over MultiMAE 

% %UNETR-based approaches like MAE-UNETR \cite{zhou2023maeunetr} demonstrate the benefits of MAE pretraining for 3D medical tasks.

% \noindent\textbf{Handling Missing Modalities and Robustness in Deep Learning for Medical Imaging}

\section{Materials and Methods}

\subsection{Dataset}

We use the 2021 edition of BraTS Glioma \cite{menze2014brats}, along with the UPENN-GBM \cite{bakas2022upenn}, UCSF-PGDM \cite{calabrese2022ucsf}, Lumiere \cite{suter2022lumiere}, Rembrandt \cite{sayah2022rembrandt}, EGD \cite{van2021erasmus}, TCGA \cite{weinstein2013tcga}, and the 2023 edition of the BraTS Metastasis dataset \cite{moawad2024bratsmets}. Patient overlaps between BraTS'21 and other sources are removed. The combined dataset is randomly split for pretraining as $3243$/$650$/$434$ (train/val/test). For out-of-distribution, external evaluation, we additionally use in-house datasets comprising $252$ glioma and $167$ brain metastasis cases.

Each downstream dataset is refined to exclude irrelevant or postoperative scans. Glioma segmentation uses $2676$/$536$/$358+252$, metastasis segmentation $567$/$114$/$76+167$, and subtype classification $1253$/$251$/$168+203$ patients, with GBM/Astro/Oligo prevalence of $80$/$13$/$7\%$ internally and $77$/$12$/$11\% $externally.

All four MRI sequences (t1, t1c, t2, fla) are skull-stripped, resampled to 1mm\textsuperscript{3} isotropic resolution, and registered to the SRI24 atlas \cite{rohlfing2010sri24}. Expert-made segmentations following the BraTS convention are available for all patients, except for the external metastasis dataset, which lacks enhancing tumor labels.

\subsection{Model Architecture and Pretraining}
We adopt the MultiMAE \cite{bachmann2022multimae} framework using a modified ViT, which encodes shared representations across multiple imaging modalities as our backbone.

Non-overlapping $16\times16\times16$ patches from each modality are projected into a shared token space via modality-specific linear adapters.
A global masking ratio is set as 75\%, with per-modality masking sampled from a Dirichlet distribution ($\alpha=1$), allowing up to $100\%$ masking of a modality while preserving the expected global rate. Unmasked tokens from all modalities, along with a \texttt{[cls]} token, form the input sequence.

Each modality is reconstructed using a dedicated output transformer. For masked patches, learnable \texttt{[mask]} tokens are inserted at their original positions to form the query sequence. As in \cite{bachmann2022multimae}, the decoders are implemented as \textit{cross-attention transformers}: queries first attend to the context via a single cross-attention layer, followed by standard self-attention layers. The modality-specific masking scheme encourages each decoder to reconstruct missing patches using both its visible tokens and the shared multi-modal context, enabling the model to capture rich cross-modal dependencies during training.

Following prior work \cite{he2022masked,bachmann2022multimae,zhou2023maeunetr}, we adopt an asymmetric encoder–decoder design. The encoder is a ViT-B/16 architecture ($12$ layers; token size: $t \in \mathbb{R}^{768}$; $12$ attention heads). The modality-specific decoders are lightweight (3 layers; $t \in \mathbb{R}^{384}$; 12 heads). Both the encoder and decoders use 3D sine-cosine positional embeddings. 

The pretraining optimizes reconstruction using an MSE loss over all tokens. To reduce redundancy from background-only regions and maintain compatibility with the patch size, we crop inputs to $160\times176\times144$ around the foreground.

% Unlike the original MultiMAE, we compute reconstruction losses over tokens, not just the masked ones, to stabilize training in the medical setting.
% \input{fig_architecture}
\begin{figure}
    \centering
    \includegraphics[width=\linewidth]{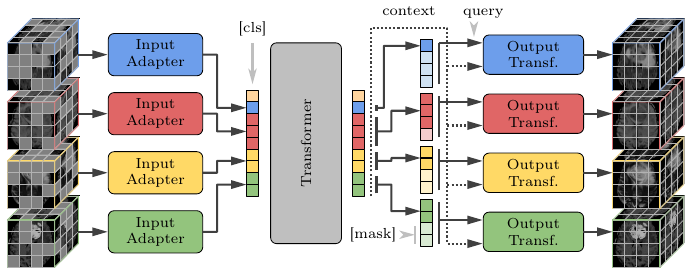}
    \caption{\textbf{Adapting the MultiMAE workflow for brain MRIs}}
    \label{fig:enter-label}
\end{figure}

\subsection{Adaptation to Downstream Tasks}

\textbf{Segmentation} We use the UNETR \cite{zhou2023maeunetr} architecture for its off-the-shelf applicability without modifications to the ViT encoder. It performs segmentation via convolutional upsampling of patch tokens from different encoder layers, similar to skip connections of U-Net \cite{ronneberger2015unet}. However, it expects a fixed-length token sequence for spatial consistency. Since each input modality is tokenized separately in our method, the sequence length is variable. We maintain a consistent length by averaging the tokens at each spatial location.

We randomly crop the MRI sequences into $128\times128\times128$ during finetuning. At test time, shape consistency is achieved through sliding window inference with a $25\%$ overlap. The model is optimized using the Dice loss.

\noindent \textbf{Classification} For classification, the decoders are replaced with a single linear layer $\mathbb{R}^{768\times3}$. To aggregate the patch tokens into a single feature vector, we average them together with the CLS token. 

We finetune the network with cross-entropy loss. The minority classes, Oligo and Astro, are oversampled to match a uniform distribution with GBM, mitigating the common class imbalance in medical imaging \cite{scholz2024imbalance}. %Sequences are cropped to $96\times96\times96$ around the tumor for a more focused view.

\subsection{Evaluation}

For each downstream task, we train a vanilla ViT-B/16 baseline that handles multi-modal MRIs through stacking along the channel dimension. Following \cite{zhou2023maeunetr}, we perform a MAE pretraining for this ViT with the same masking ratio and decoder configuration as in their work (8 layers; token size: $t \in \mathbb{R}^{384}$; 12 heads).

We monitor the effectiveness of pretraining and the thus achieved MRI generation via PSNR and SSIM. Segmentation is evaluated using the Dice score on enhancing tumor (ET), tumor core (TC), and whole tumor (WT), per BraTS guidelines. For classification tasks, we report the multi-class accuracy, F1 score, and Matthew's correlation coefficient (MCC).

All four MRI sequences are provided during finetuning. At test time, we evaluate five scenarios: one with all inputs and four with a single modality removed. Tables in \autoref{sec:results} depict these scenarios. For our model, the missing modality is simply excluded, while for the vanilla ViT, it is filled with background intensities due to the fixed input shape. No masking is applied during finetuning, making this a truly \textit{unprepared} setting for evaluating robustness to unseen input configurations.

\subsection{Training Setup}

All pretrainings run for 1200 epochs with a batch size of $16$. Downstream finetunings and their from-scratch counterparts are performed for 100 epochs. Classification uses the same batch size, while segmentation is limited to two due to memory constraints (NVIDIA A40 GPUs). We use the AdamW optimizer with a learning rate of $1\mathrm{e}{-4}$ and $0.05$ weight decay. During MAE pretrainings, the learning rate is rescaled by $0.1$ on a plateau of $50$ epochs. For downstream tasks, we apply cosine decay with $40$ warmup epochs. A gradient clipping by $0.5$ is applied in all settings.

\section{Results}\label{sec:results}

\subsection{Missing MRI Generation}
The multi-task MAE pretraining naturally enables image generation, treating the missing modality as a fully masked input. The encoder-decoder structure leverages its cross-modal capabilities to generate the missing input based on the three present inputs.

\begin{figure}
    \centering

    \includegraphics[width=0.75\textwidth]{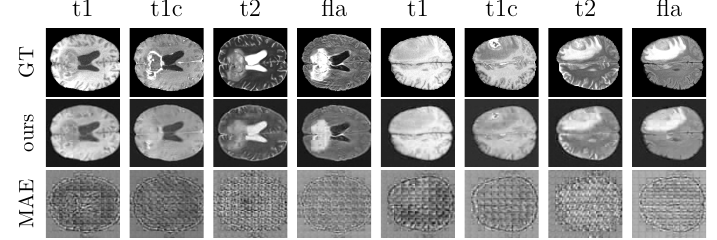}
    \caption{\textbf{Reconstruction of a missing modality}. The depicted modality was fully masked (for MAE: replaced with background), and then reconstructed using the remaining inputs.}
    \label{fig:reconstruction}
\end{figure}

As shown in \autoref{tab:dropout_reconstruction} and \autoref{fig:reconstruction}, the model produces anatomically realistic outputs, though some blurring and loss of fine detail remain. In contrast, the vanilla ViT fails to achieve comparable reconstruction quality, even after pretraining.

\input{tab_dropout-reconstruction}

\subsection{Segmentation}
% - from table 2 and 3: with pretraining, our model is better in almost all cases compared to the MAE-ViT, both in internal and external data
% - without even needs full finetuning and can achieve high performance with frozen encoder 
% - table 3 further shows how beneficial our pretraining is, with the jump in results. maybe also a limitation to use it from scratch (randomly initialized)
% - our method is much more robust against missing modalities. the pattern in missing modality experiments aligns with clinical knowledge: T1 and T1CE are more predictive of ET and TC, whereas WT is dependent on T2 variant. our model learns to look for the relevant input and maintain performance, whereas ViT fails more drastically in all targets

As shown in \autoref{tab:table2} and \autoref{tab:table3}, our pretrained model consistently outperforms the MAE-ViT baseline, both on internal and external datasets. Despite using a frozen encoder in finetuning for simplicity, it achieves on-par performance, indicating the effectiveness of the learned representations. On the more underexplored task of metastasis segmentation, \autoref{tab:table3} highlights the value of our pretraining; there is substantial improvement compared to training \textit{from scratch} (randomly initialized). 

In missing modality experiments, our method proves far more robust than ViT. Performance patterns align with clinical expectations, e.g., t1 and t1c are more predictive for ET and TC, while WT depends more on FLAIR, showing that our model learns to leverage the most relevant inputs, whereas ViT suffers higher performance drops overall, even when key modalities are not missing.
\input{tab_gbm-seg_internal-external}
\input{tab_met-seg_internal-external}

\subsection{Classification}

In the multi-class glioma subtyping task, a similar trend to segmentation is observed: our pretrained model consistently outperforms the MAE-ViT baseline across both internal (\autoref{tab:table4}) and external (\autoref{tab:table5}) datasets. It achieves strong performance, with MCC scores reaching up to $0.70$. Remarkably, this performance is largely retained even when one modality is missing, whereas the ViT baseline shows a more substantial drop. A moderate decline is only observed in the absence of t1c, which is expected due to its critical role in characterizing tumor enhancement. 

In contrast to segmentation, freezing the encoder during finetuning is less effective for classification, likely because a single linear layer is insufficient to align the learned representations with the task. This limitation is also evident for the vanilla ViT. However, full finetuning, which is more feasible here due to the model size, boosts performance.

\input{tab_gbm-subtype_internal}
\input{tab_gbm-subtype_external_full}

\section{Conclusion}

In this work, we present a MultiMAE pretraining framework for 3D brain MRI using masked modeling across multiple sequences and modality-specific encoding. Unlike early-fusion methods that treat all modalities as a single block, our approach learns distinct yet complementary representations per sequence, enhancing robustness to missing inputs common in clinical practice.

Our pretrained model consistently outperforms a baseline MAE-ViT in both segmentation and classification, showing strong performance on internal data and good generalization externally. However, a domain gap persists, reflecting the common challenge of adapting across clinical sites in medical imaging.

Particularly notable is the model's resilience to missing modalities: it continues to perform competitively even when an entire input sequence is absent, and is capable of extracting the information through learned cross-modal relationships. This is further evidenced by the model’s strong ability to synthesize a missing modality from the remaining three. Still, the reconstructions show some blurring and loss of fine details, likely due to the limitations of pixel-wise MSE loss. Future work could address this by exploring perceptual or feature-level loss functions to enhance reconstruction fidelity.

Nonetheless, our approach shows greater dependence on pretraining compared to a vanilla ViT, with randomly initialized versions yielding weaker performance in some cases. This likely suggests that the added complexity of cross-modal integration demands more data or longer training to converge effectively. While pretraining requires considerable compute, it is performed only once and with minimal additional cost, can then be flexibly adapted to diverse tasks and resource-constrained environments.

Finally, the downstream performance is ultimately bounded by the used decoders; future gains may come from more advanced decoding strategies.

In conclusion, we present a simple yet effective reformulation of MAE pretraining paradigm that enables more elegant handling of multi-modal brain MRI data.

\begin{credits}
\subsubsection{\ackname} 
This study was supported by the DFG, grant \#504320104.

\subsubsection{\discintname}
The authors have no competing interests to declare that are relevant to the content of this article.
\end{credits}


% \newpage
\bibliographystyle{splncs04}
% \bibliography{bibliography}

\input{main.bbl}
\end{document}

%% file: tab_dropout-reconstruction.tex
\begin{table}[!ht]
    \centering
    \caption{MRI sequence generation performance. Each column reports the reconstruction quality (PSNR and SSIM) for the "missing" MRI sequence, generated from the remaining three modalities. A ViT pretrained with MAE is shown for comparison.}
    \label{tab:dropout_reconstruction}
    \setlength{\tabcolsep}{4pt}      % Adjust  column spacing
    \begin{tabular}{
        @{\hspace{12pt}} l
        @{\hspace{12pt}} cccc
        @{\hspace{12pt}} cccc
    }
        \toprule
        \multicolumn{1}{c@{\hspace{12pt}}}{\textbf{Backbone}} & \multicolumn{4}{c@{\hspace{12pt}}}{\textbf{PSNR}} & \multicolumn{4}{c@{\hspace{12pt}}}{\textbf{SSIM}} \\
         & t1 & t1c & t2 & fla & t1 & t1c & t2 & fla \\ 
        \midrule
        ViT & 7.36 & 8.18 & 10.36 & 8.01 & 0.52 & 0.54 & 0.61 & 0.55 \\
        ours & \textbf{21.58} & \textbf{20.56} & \textbf{21.01} & \textbf{20.74} & \textbf{0.76} & \textbf{0.71} & \textbf{0.71} & \textbf{0.71} \\
        \bottomrule
    \end{tabular}
\end{table}

%% file: tab_gbm-seg_internal-external.tex
\begin{table}[!ht]
    % \caption{\textbf{Glioma Segmentation (Internal + External)}}
    \caption{\textbf{Glioma segmentation (Internal + External)}. Dice scores for ViT and our pretrained model on internal and external test sets. We report performance for tumor core (TC), enhancing tumor (ET), and whole tumor (WT) under three training regimes: scratch, frozen (internal), and frozen (external).}
    
    \centering
    % \caption{Glioma Segmentation Internal + External}
    
    \label{tab:table2}
    \renewcommand{\arraystretch}{0.7} % Adjust row height for better readability
    \begin{tabular}{
        cccc
        @{\hspace{6pt}} l
        % @{\hspace{24pt}} c
        @{\hspace{6pt}} ccc
        @{\hspace{6pt}} ccc
        @{\hspace{6pt}} ccc
    }
        \toprule
         \multicolumn{4}{c@{\hspace{6pt}}}{\textbf{Modalities}} & \multicolumn{1}{l@{\hspace{6pt}}}{\textbf{Enc.}} & \multicolumn{3}{c@{\hspace{6pt}}}{\textbf{TC}} & \multicolumn{3}{c@{\hspace{6pt}}}{\textbf{ET}} & \multicolumn{3}{c@{\hspace{6pt}}}{\textbf{WT}} \\
        % T1 & T1C & T2 & FLA & & scr. & frozen & full & scr. & frozen & full & scr. & frozen & full \\ 

        \cmidrule(lr){6-8} \cmidrule(lr){9-11} \cmidrule(lr){12-14}
        \multirow{2}{*}{t1} & \multirow{2}{*}{t1c} & \multirow{2}{*}{t2} & \multirow{2}{*}{fla} & & \multicolumn{1}{c@{\hspace{6pt}}}{\textbf{scr.}} & \multicolumn{2}{c@{\hspace{6pt}}}{\textbf{frozen}} & \multicolumn{1}{c@{\hspace{6pt}}}{\textbf{scr.}} & \multicolumn{2}{c@{\hspace{6pt}}}{\textbf{frozen}} & \multicolumn{1}{c@{\hspace{6pt}}}{\textbf{scr.}} & \multicolumn{2}{c@{\hspace{6pt}}}{\textbf{frozen}} \\ 
        \cmidrule(lr){6-6} \cmidrule(lr){7-8} \cmidrule(lr){9-9} \cmidrule(lr){10-11} \cmidrule(lr){12-12} \cmidrule(lr){13-14}
        &&&&& int. & int. & ext. & int. & int. & ext. & int. & int. & ext. \\
        \midrule
        \multirow{2}{*}{\ding{51}} & \multirow{2}{*}{\ding{51}} & \multirow{2}{*}{\ding{51}} & \multirow{2}{*}{\ding{51}} & ViT & 73.75 & 74.30 & \textbf{67.90} & 71.00 & 72.33 & 66.43 & 81.48 & 80.40 & 74.73 \\
         & & &  & ours & 72.36 & \textbf{75.15} & 65.75 & 71.07 & \textbf{73.00} & \textbf{66.84} & 80.73 & \textbf{82.78} & \textbf{78.80} \\
        \midrule
        \multirow{2}{*}{\ding{55}} & \multirow{2}{*}{\ding{51}} & \multirow{2}{*}{\ding{51}} & \multirow{2}{*}{\ding{51}} & ViT & 22.07 & 47.91 & 30.93 & 21.16 & 45.76 & 38.19 & 25.66 & 47.85 & 48.61 \\
         & & &  & ours & 8.11 & \textbf{51.55} & \textbf{45.00} & 6.95 & \textbf{48.96} & \textbf{49.80} & 19.44 & \textbf{79.03} & \textbf{76.54} \\
        \midrule
        \multirow{2}{*}{\ding{51}} & \multirow{2}{*}{\ding{55}} & \multirow{2}{*}{\ding{51}} & \multirow{2}{*}{\ding{51}} & ViT & 25.32 & 21.55 & 25.07 & 6.99 & 7.23 & 12.56 & 41.77 & 64.98 & 55.91 \\
         & & &  & ours & 13.64 & \textbf{37.85} & \textbf{32.90} & \textbf{7.46} & \textbf{7.46} & \textbf{12.81} & 53.58 & \textbf{80.08} & \textbf{76.59} \\
        \midrule
        \multirow{2}{*}{\ding{51}} & \multirow{2}{*}{\ding{51}} & \multirow{2}{*}{\ding{55}} & \multirow{2}{*}{\ding{51}} & ViT & 50.04 & 64.86 & 49.34 & 60.72 & 64.98 & 50.98 & 55.71 & \textbf{64.00} & \textbf{44.64} \\
         & & &  & ours & 28.81 & \textbf{66.72} & \textbf{60.14} & 57.07 & \textbf{69.30} & \textbf{62.75} & 25.83 & 53.55 & 36.34 \\
        \midrule
        \multirow{2}{*}{\ding{51}} & \multirow{2}{*}{\ding{51}} & \multirow{2}{*}{\ding{51}} & \multirow{2}{*}{\ding{55}} & ViT & 0.74 & 14.42 & 18.96 & 2.03 & 17.00 & 22.65 & 0.05 & 5.29 & 4.31 \\
         & & &  & ours & 16.88 & \textbf{38.91} & \textbf{30.95} & 15.50 & \textbf{37.11} & \textbf{34.96} & 7.63 & \textbf{16.69} & \textbf{10.12} \\
        \bottomrule
    \end{tabular}
\end{table}

%% file: tab_met-seg_internal-external.tex
\begin{table}[!ht]
    \caption{\textbf{Metastasis Segmentation (Internal + External)} Enhancing tumor (ET) labels are not available for the external test set.}

    \centering
        \label{tab:table3}
    \renewcommand{\arraystretch}{0.7} % Adjust row height for better readability
    \begin{tabular}{
        cccc
        @{\hspace{6pt}} l
        % @{\hspace{24pt}} c
        @{\hspace{6pt}} ccc
        @{\hspace{6pt}} cc
        @{\hspace{6pt}} ccc
    }
        \toprule
         \multicolumn{4}{c@{\hspace{6pt}}}{\textbf{Modalities}} & \multicolumn{1}{l@{\hspace{6pt}}}{\textbf{Enc.}} & \multicolumn{3}{c@{\hspace{6pt}}}{\textbf{TC}} & \multicolumn{2}{c@{\hspace{6pt}}}{\textbf{ET}} & \multicolumn{3}{c@{\hspace{6pt}}}{\textbf{WT}} \\
        % T1 & T1C & T2 & FLA & & scratch & frozen & full & scratch & frozen & full & scratch & frozen & full \\ 

        \cmidrule(lr){6-8} \cmidrule(lr){9-10} \cmidrule(lr){11-13}
        \multirow{2}{*}{t1} & \multirow{2}{*}{t1c} & \multirow{2}{*}{t2} & \multirow{2}{*}{fla} & & \multicolumn{1}{c@{\hspace{6pt}}}{\textbf{scr.}} & \multicolumn{2}{c@{\hspace{6pt}}}{\textbf{frozen}} & \multicolumn{1}{c@{\hspace{6pt}}}{\textbf{scr.}} & \multicolumn{1}{c@{\hspace{6pt}}}{\textbf{frozen}} & \multicolumn{1}{c@{\hspace{6pt}}}{\textbf{scr.}} & \multicolumn{2}{c@{\hspace{6pt}}}{\textbf{frozen}} \\ 
        \cmidrule(lr){6-6} \cmidrule(lr){7-8} \cmidrule(lr){9-9} \cmidrule(lr){10-10} \cmidrule(lr){11-11} \cmidrule(lr){12-13}
        &&&&& int. & int. & ext. & int. & int. & int. & int. & ext. \\
        \midrule
        \multirow{2}{*}{\ding{51}} & \multirow{2}{*}{\ding{51}} & \multirow{2}{*}{\ding{51}} & \multirow{2}{*}{\ding{51}} & ViT & 38.40 & 42.19 & 33.39 & 40.87 & 37.58 & 53.76 & 54.61 & 48.55 \\
         & & &  & ours & 15.28 & \textbf{58.19} & \textbf{51.67} & 16.76 & \textbf{63.48} & 31.71 & \textbf{75.07} & \textbf{63.24} \\
        \midrule
        \multirow{2}{*}{\ding{55}} & \multirow{2}{*}{\ding{51}} & \multirow{2}{*}{\ding{51}} & \multirow{2}{*}{\ding{51}} & ViT & 23.71 & 16.98 & 5.61 & 18.34 & 8.55 & 25.74 & 22.91 & 14.63 \\
         & & &  & ours & 1.17 & \textbf{46.43} & \textbf{31.22} & 0.73 & \textbf{35.08} & 3.19 & \textbf{66.92} & \textbf{58.45} \\
        \midrule
        \multirow{2}{*}{\ding{51}} & \multirow{2}{*}{\ding{55}} & \multirow{2}{*}{\ding{51}} & \multirow{2}{*}{\ding{51}} & ViT & 0.71 & 2.33 & 4.86 & 1.55 & 6.61 & 24.93 & 58.70 & 46.40 \\
         & & &  & ours & \textbf{2.98} & 1.72 & \textbf{6.29} & 1.49 & \textbf{7.46} & 24.24 & \textbf{72.89} & \textbf{62.93} \\
        \midrule
        \multirow{2}{*}{\ding{51}} & \multirow{2}{*}{\ding{51}} & \multirow{2}{*}{\ding{55}} & \multirow{2}{*}{\ding{51}} & ViT & 8.57 & 13.93 & 21.69 & 14.70 & 4.72 & 3.34 & 18.05 & \textbf{33.68} \\
         & & &  & ours & 8.41 & \textbf{38.21} & \textbf{35.87} & 9.50 & \textbf{47.39} & 9.93 & \textbf{21.17} & 12.13 \\
        \midrule
        \multirow{2}{*}{\ding{51}} & \multirow{2}{*}{\ding{51}} & \multirow{2}{*}{\ding{51}} & \multirow{2}{*}{\ding{55}} & ViT & 2.96 & 1.82 & 5.08 & 3.51 & 5.51 & 5.16 & 0.10 & 0.04 \\
         & & &  & ours & 0.88 & \textbf{30.58} & \textbf{17.59} & 1.06 & \textbf{31.64} & 0.12 & \textbf{8.77} & \textbf{3.98} \\
        \bottomrule
    \end{tabular}
\end{table}

%% file: tab_gbm-subtype_internal.tex
\begin{table}[h!]
    \centering
 
    \renewcommand{\arraystretch}{0.7} % Adjust row height for better readability
    \caption{\textbf{Glioma Subtype Classification (Internal)} All metrics are computed for the multi-class setting.} %\textit{no} indicates scr. pretraining, \textit{frozen} is finetuning with frozen encoder, and \textit{full} is complete finetuning after respective pretraining }
   \label{tab:table4}
    
    \begin{tabular}{
        cccc
        @{\hspace{6pt}} l 
        % @{\hspace{24pt}} c
        @{\hspace{6pt}} ccc
        @{\hspace{6pt}} ccc
        @{\hspace{6pt}} ccc
    }
        \toprule
         \multicolumn{4}{c@{\hspace{6pt}}}{\textbf{Modalities}} & \multicolumn{1}{l@{\hspace{6pt}}}{\textbf{Enc.}} & \multicolumn{3}{c@{\hspace{6pt}}}{\textbf{Accuracy}} & \multicolumn{3}{c@{\hspace{6pt}}}{\textbf{F1}} & \multicolumn{3}{c@{\hspace{6pt}}}{\textbf{MCC}} \\
        % t1 & t1c & t2 & fla & & scratch & frozen & full & scratch & frozen & full & scratch & frozen & full \\ 
        t1 & t1c & t2 & fla & & scr. & frozen & full & scr. & frozen & full & scr. & frozen & full \\ 
        \midrule
        \multirow{2}{*}{\ding{51}} & \multirow{2}{*}{\ding{51}} & \multirow{2}{*}{\ding{51}} & \multirow{2}{*}{\ding{51}} & ViT & 0.85 & 0.67 & 0.89 & 0.67 & 0.49 & 0.63 & 0.59 & 0.39 & 0.66 \\
         & & &  & ours & 0.73 & 0.68 & \textbf{0.90} & 0.54 & 0.49 & \textbf{0.73} & 0.43 & 0.38 & \textbf{0.69} \\
        \midrule
        \multirow{2}{*}{\ding{55}} & \multirow{2}{*}{\ding{51}} & \multirow{2}{*}{\ding{51}} & \multirow{2}{*}{\ding{51}} & ViT & 0.51 & 0.13 & 0.41 & 0.35 & 0.08 & 0.26 & 0.15 & 0.00 & 0.03 \\
         & & &  & ours & 0.61 & 0.71 & \textbf{0.89} & 0.46 & 0.49 & \textbf{0.70} & 0.31 & 0.36 & \textbf{0.66} \\
        \midrule
        \multirow{2}{*}{\ding{51}} & \multirow{2}{*}{\ding{55}} & \multirow{2}{*}{\ding{51}} & \multirow{2}{*}{\ding{51}} & ViT & 0.25 & 0.70 & 0.06 & 0.17 & 0.41 & 0.04 & 0.12 & 0.22 & -0.10 \\
         & & &  & ours & 0.67 & 0.47 & \textbf{0.79} & 0.46 & 0.36 & \textbf{0.66} & 0.27 & 0.20 & \textbf{0.47} \\
        \midrule
        \multirow{2}{*}{\ding{51}} & \multirow{2}{*}{\ding{51}} & \multirow{2}{*}{\ding{55}} & \multirow{2}{*}{\ding{51}} & ViT & 0.58 & 0.13 & 0.75 & 0.37 & 0.08 & 0.43 & 0.36 & 0.0 & 0.36 \\
         & & &  & ours & 0.5 & 0.76 & \textbf{0.89} & 0.40 & 0.53 & \textbf{0.66} & 0.31 & 0.40 & \textbf{0.66} \\
        \midrule
        \multirow{2}{*}{\ding{51}} & \multirow{2}{*}{\ding{51}} & \multirow{2}{*}{\ding{51}} & \multirow{2}{*}{\ding{55}} & ViT & 0.45 & 0.78 & 0.81 & 0.30 & 0.41 & 0.46 & 0.13 & 0.17 & 0.37 \\
         & & &  & ours & 0.81 & 0.64 & \textbf{0.91} & 0.50 & 0.46 & \textbf{0.74} & 0.41 & 0.35 & \textbf{0.71} \\
    \bottomrule
    \end{tabular}
\end{table}

%% file: tab_gbm-subtype_external_full.tex
\begin{table}[h!]
    \centering
    \caption{\textbf{Glioma Subtype Classification (External)} Results are shown only for the fully finetuned model.}
    \label{tab:table5}
    \renewcommand{\arraystretch}{0.7}

    \begin{tabular}{
        c
        @{\hspace{8pt}} cc
        @{\hspace{8pt}} cc
        @{\hspace{8pt}} cc
        @{\hspace{8pt}} cc
        @{\hspace{8pt}} cc
    }
        \toprule
        \textbf{Metric} 
        & \multicolumn{2}{c}{\textbf{all}} 
        & \multicolumn{2}{c}{\textbf{no-t1}} 
        & \multicolumn{2}{c}{\textbf{no-t1c}} 
        & \multicolumn{2}{c}{\textbf{no-t2}} 
        & \multicolumn{2}{c}{\textbf{no-fla}} \\
        
        \cmidrule(lr){2-3}
        \cmidrule(lr){4-5}
        \cmidrule(lr){6-7}
        \cmidrule(lr){8-9}
        \cmidrule(lr){10-11}
        & ViT & ours & ViT & ours & ViT & ours & ViT & ours & ViT & ours \\
        \midrule
        \textbf{Accuracy} & 0.78 & \textbf{0.79} & 0.37 & \textbf{0.80} & 0.12 & \textbf{0.49} & 0.73 & \textbf{0.79} & 0.75 & \textbf{0.78} \\
        \textbf{F1} & 0.47 & \textbf{0.48} & 0.25 & \textbf{0.46} & 0.10 & \textbf{0.36} & 0.43 & \textbf{0.48} & 0.36 & \textbf{0.46} \\
        \textbf{MCC} & \textbf{0.39} & 0.36 & 0.11 & \textbf{0.38} & 0.06 & \textbf{0.12} & 0.36 & \textbf{0.39} & 0.15 & \textbf{0.33} \\
        \bottomrule
    \end{tabular}
\end{table}